\documentclass{article}

\usepackage{PRIMEarxiv}

\usepackage[utf8]{inputenc} 
\usepackage[T1]{fontenc}    
\usepackage{hyperref}       
\usepackage{url}            
\usepackage{booktabs}       
\usepackage{amsfonts}       
\usepackage{nicefrac}       
\usepackage{microtype}      
\usepackage{lipsum}
\usepackage{fancyhdr}       
\usepackage{graphicx}       
\graphicspath{{media/}}     

\title{Short-term Precipitation Forecasting in The Netherlands: An Application of Convolutional LSTM neural networks to  weather radar data}
\author{Petros Demetrakopoulos\\ 
Eindhoven University of Technology\\
\texttt{\href{mailto:p.s.dimitrakopoulos@student.tue.nl}{p.s.dimitrakopoulos@student.tue.nl}}}
\date{}

\begin{document}
\maketitle

\begin{abstract}
This work addresses the challenge of short-term precipitation forecasting by applying Convolutional Long Short-Term Memory (ConvLSTM) neural networks to weather radar data from the Royal Netherlands Meteorological Institute (KNMI). The research exploits the combination of Convolutional Neural Networks (CNNs) layers for spatial pattern recognition and LSTM network layers for modelling temporal sequences, integrating these strengths into a ConvLSTM architecture. The model was trained and validated on weather radar data from the Netherlands. The model is an autoencoder consisting of nine layers, uniquely combining convolutional operations with LSTM’s temporal processing, enabling it to capture the movement and intensity of precipitation systems. The training set comprised of sequences of radar images, with the model being tasked to predict precipitation patterns 1.5 hours ahead using the preceding data. Results indicate high accuracy in predicting the direction and intensity of precipitation movements. The findings of this study underscore the significant potential of ConvLSTM networks in meteorological forecasting, particularly in regions with complex weather patterns. It contributes to the field by offering a more accurate, data-driven approach to weather prediction, highlighting the broader applicability of ConvLSTM networks in meteorological tasks.
\end{abstract}

\keywords{Machine learning \and Deep learning \and Precipitation forecasting \and Weather forecasting}

\section{Introduction}
Meteorological forecasting, which is a critical aspect of climate science, has seen significant advancements with the integration of machine learning (ML) techniques. This research targets the complex task of short-term precipitation forecasting, also known as "nowcasting". The data used to train and validate the model come from two weather radars monitoring precipitation in The Netherlands \cite{knmiPrecipitationRadar}, a country characterized by a humid and precipitation-intense climate. The study employs Convolutional LSTM (ConvLSTM) neural networks to model these weather patterns, aiming to enhance the accuracy of short-term predictions.

\section{Literature Review and related work}
A comprehensive review of existing literature reveals a growing interest in applying deep learning techniques for weather forecasting purposes. Studies have particularly noted the efficacy of Convolutional Neural Networks (CNNs) in spatial pattern recognition and Long Short-Term Memory (LSTM) networks in modelling temporal sequences. The fusion of these two approaches in ConvLSTM architectures has shown promising results in various fields, including video frame prediction and climate modelling. The most noteworthy related publication is the one titled "Convolutional LSTM network: A machine learning approach for precipitation nowcasting" by \cite{DBLP:journals/corr/ShiCWYWW15} X. Shi et. al. published in 2015. Other published works have also proposed Generative Adversarial Network architectures with a worth-mentioning paper by L. Xu et. al. \cite{rs14235948} published in 2022. 

\subsection{Data Acquisition and Preprocessing}\label{sec:data}
\begin{figure}
    \centering
    \includegraphics[scale=0.4]{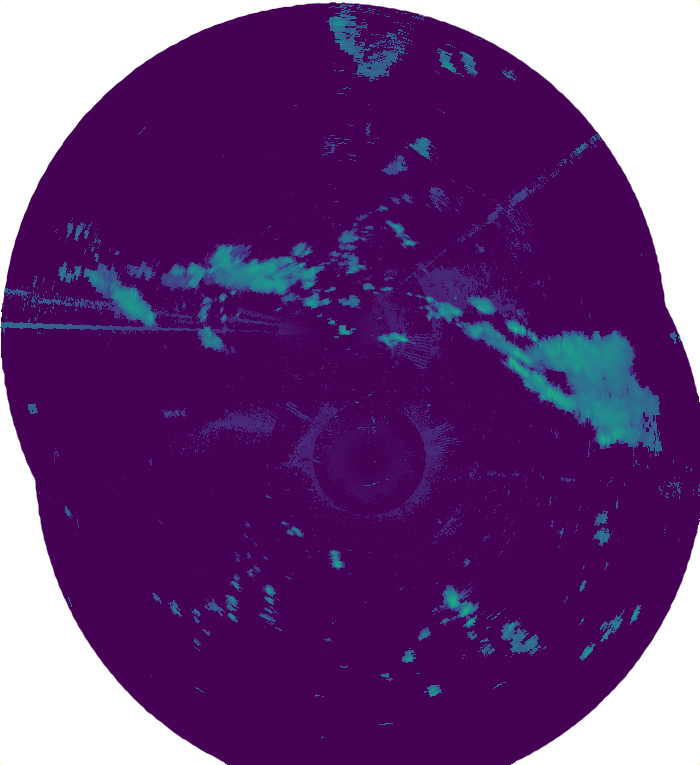}
    \caption{Example of a radar image after preprocessing. The darker regions indicate lower reflectivity values while the lighter coloured regions indicate higher reflectivity.}
    \label{fig:example}
\end{figure}

Data from two Dutch weather radars, the first being located in Herwijnen and the second one in Den Helder were captured and converted into a visual format. In an effort to avoid overly technical explanations of weather radar mechanisms that are out of the scope of this work, it is known that weather radars emit a signal which interacts with various forms of precipitation (including rain, snow, and hail). This interaction results in the reflection of the emitted signal back to the radar. The magnitude of this reflected signal, commonly referred to as 'reflectivity', is measured in decibels (dB). It is generally accepted that reflectivity bears a direct, albeit approximate, correlation to the intensity of the precipitation at the point of signal reflection \cite{RINEHART1978}, \cite{RADARREFLECTIVITYPROFILESINTHUNDERSTORMS}. Subsequently, this reflectivity data is transformed into a visual representation through the application of a colour scale corresponding to the signal intensity levels. Data were fetched via the Royal Netherlands Meteorological Institute's  (KNMI) open data public API \cite{knmiPrecipitationRadar}. The default colour scale suggested and used by KNMI is 'Viridis', which denotes lower signal intensities in hues of purple and dark blue, and higher intensities in yellow. The study faced limitations due to KNMI's weather radar API constraints, allowing to fetch only a limited number of images per hour. The preprocessing involved downsizing the images from their original high-resolution format ($765\times760$) to a more manageable size, specifically $344\times315$, facilitating efficient model training without significant loss of information. An example of how a radar image is after the preprocessing steps mentioned above is presented in Figure \ref{fig:example}

\subsection{Methodology}
It is well-established from previous research that Convolutional Neural Networks (CNNs) \cite{lecun-gradientbased-learning-applied-1998} exhibit proficient performance in tasks involving the analysis and identification of distinct features and shapes within images. Conversely, Long Short-Term Memory (LSTM) \cite{HochSchm97} networks are notably effective in tasks encompassing temporal dynamics, such as time series predictions, and in processing sequential data, which includes sequences of images or signals over specific time frames. This effectiveness is attributed to their inherent capability to learn and retain long-term dependencies in data sets. The core of the methodology lies in the development and implementation of a ConvLSTM neural network originally proposed by Shi et. al.\cite{DBLP:journals/corr/ShiCWYWW15}. This neural network architecture is designed to process spatiotemporal data effectively, capturing both the spatial features of precipitation patterns and their temporal evolution, the internal details and design are covered in the following subsection and in the original paper \cite{DBLP:journals/corr/ShiCWYWW15}. The GitHub repository containing the implementation of the model described in the coming section as well as the training and validation datasets is located in the following link \url{https://github.com/petrosDemetrakopoulos/LSTM-radar-precipitation-forecast} \cite{githubGitHubPetrosDemetrakopoulosLSTMradarprecipitationforecast}.

\subsection{Model Architecture:}
\begin{figure}
    \centering
    \includegraphics[scale=0.45]{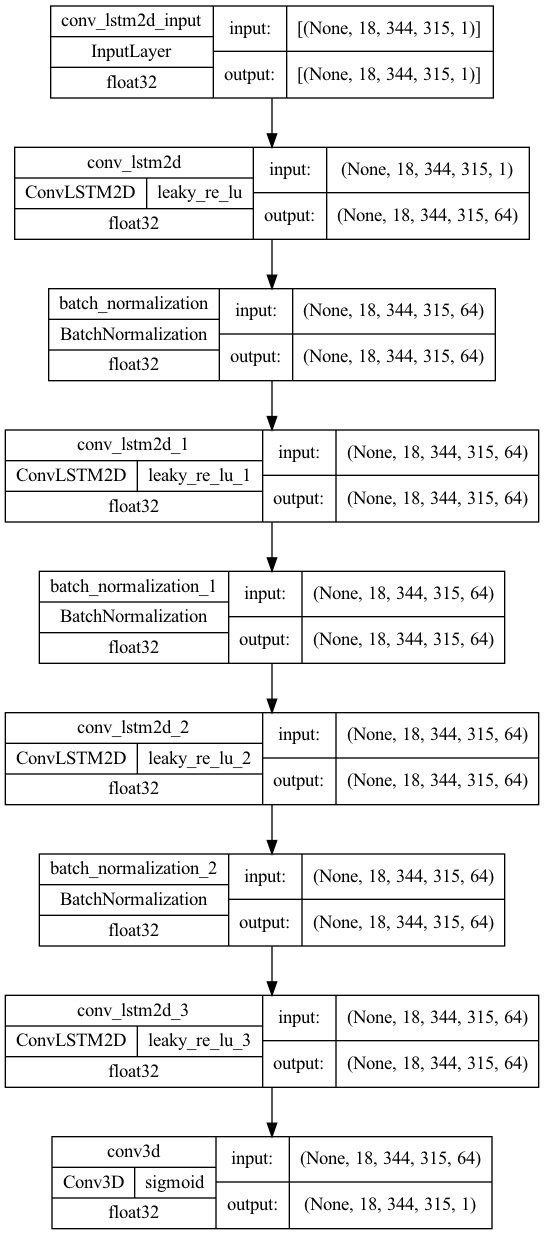}
    \caption{Neural Network Architecture}
    \label{fig:model}
\end{figure}

The proposed neural network is an autoencoder consisting of nine layers: an input layer, seven hidden layers (alternating between ConvLSTM2D and Batch Normalization layers), and an output layer. The ConvLSTM2D layers, combine convolutional operations with LSTM's temporal processing. This trait allows the model to capture temporal relationships between different frames/images produced from weather radar signals and thus learn the ways that precipitation systems and clouds move. Batch Normalization layers are employed to stabilize learning by normalizing layer inputs. LeakyReLU is chosen as the activation function to address issues of sparse gradients and avoid the vanishing gradient problem \cite{maasrectifier}. The network architecture is presented in a schematic way in Figure \ref{fig:model}. The model was implemented in TensorFlow \cite{tensorflow2015-whitepaper} using the Keras API \cite{chollet2015keras}.
\subsection{Training and Validation:}
The dataset for each training instance comprised 36 sequential raw radar files, representing a temporal span of three hours at five-minute intervals. These data points were split into two segments. The initial 18 frames functioned as the input features ($X$), while the objective of the neural network was to predict the subsequent 18 frames ($y$), using the first 18 frames as a basis. In meteorological forecasting terms, this equates to predicting the location and intensity of precipitation for the forthcoming 1.5 hours, given the precipitation data from the preceding 1.5 hours. The model was trained on a dataset containing 400 sequences of weather radar images retrieved as mentioned in subsection \ref{sec:data}. Each image contained the composite signal of both weather radars. To ensure that the model is able to generalize well and learn seasonal patterns that may affect precipitation, training data were uniformly distributed across different months from October 2019 to December 2022.  The training process focused on minimizing the Binary Cross-Entropy loss function using the Adadelta optimizer \cite{DBLP:journals/corr/abs-1212-5701}, chosen for its efficiency in handling high-dimensional data. The 400 sequences of 36 sequential radar images each were used for training and 50 different sequences were used for validation. The validation sequences were not included in the training dataset as the relevant ML validation bibliography suggests and there were also data from all months of the year to account for differences between different seasons. In addition, to ensure certain diversity between the training and validation datasets, the validation dataset mostly contained weather radar data from 2023 while the training dataset mostly contained data from 2022. The model was trained for 25 epochs. Due to the complex design and computationally heavy operations needed, the Google Colab \cite{Bisong2019} cloud model training service was used for the training of the neural network model. Indicatively, each epoch takes roughly 6 hours and 20 minutes to be completed in an M1 Max CPU with 32 GBs of RAM.

\section{Results and Analysis}
\begin{figure}
    \centering
    \includegraphics[scale=0.5]{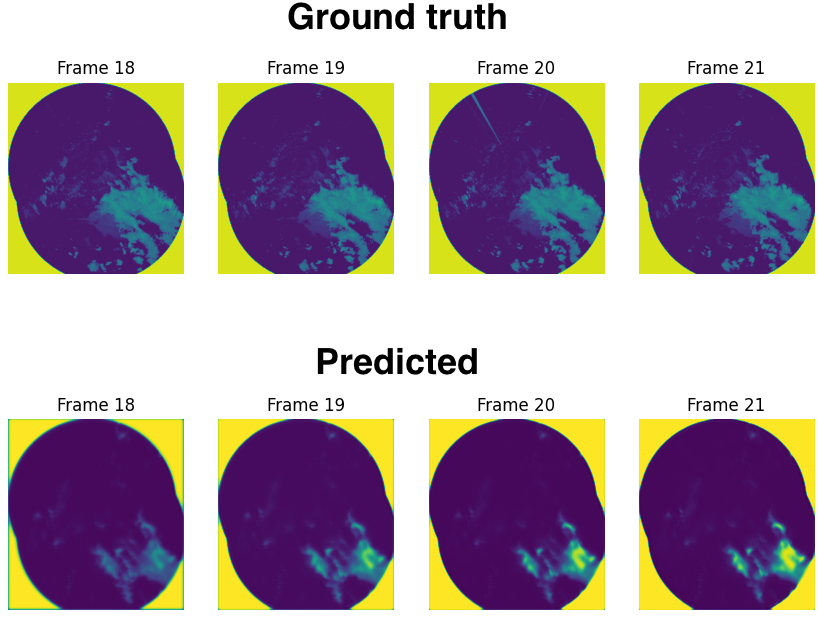}
    \caption{Example of the 4 first frames of a predicted data point side-by-side with the Ground Truth}
    \label{fig:results}
\end{figure}

The model's performance was evaluated based on its ability to predict precipitation patterns 1.5 hours ahead of the time that input radar signals were given, compared against the actual radar data captured 1.5 hours after the time that the input signals were given to the model. By abstractly looking at the task, one could claim that the problem is a regression task and thus the most appropriate metric to measure performance is the Root Mean Square Error (RMSE). The RMSE metric was calculated between the 18 ground truth (actual radar images) frames coming after the first 18 frames used as the input for the model and the frames predicted by the model. The results indicated a high degree of accuracy in predicting both the direction and intensity of precipitation movements. Due to the very heavy nature of the model training, it was only possible to perform performance experiments and report metrics for the holdout (validation) dataset. The model achieved an RMSE of 0.08246 in the validation dataset comprising of 50 sequences. An example of the predictions (4 first predicted frames) performed by the model for a given data point (18 sequential radar images) from the holdout dataset presented side-by-side with the related ground truth radar images can be seen in Figure \ref{fig:results}.  A GIF animation showing ground truth and predicted frames side-by-side and how they evolve over time is also available on the GitHub repository hosting the project \cite{githubGitHubPetrosDemetrakopoulosLSTMradarprecipitationforecast}. It can be observed that although predicted frames are somewhat noisy, the model has properly captured the general direction and intensity of precipitation systems.
\section{Discussion}
The model appears capable of predicting how precipitation patterns will move in the short term given how they moved the prior 90 minutes. The fact that the model is trained with data from a specific area makes it very accurate for the prediction of precipitation in a very specific region (i.e. a country). and most probably will not generalize well to predicting precipitation movement in other areas. Another limitation that the model has is its high complexity. The model's complexity makes the training and updating process of the model very challenging and resource-intensive. In case climate change changes the movement and intensity of precipitation patterns in the long term (concept drift), the model will need to be retrained to provide accurate predictions. Apart from these limitations, the model provides very accurate predictions based on the results reported in the previous section which are comparable to other relevant studies (Shi et. al. \cite{DBLP:journals/corr/ShiCWYWW15}, Xu et. al. \cite{rs14235948})
\section{Conclusion}
The study demonstrates the significant potential of ConvLSTM networks in short-term precipitation forecasting, combining spatial and temporal data processing capabilities. This research contributes to the field of meteorology by providing a more accurate, data-driven approach to weather prediction, particularly in regions with complex weather patterns like The Netherlands. The study also highlights the importance of ML-based weather prediction models targeted at smaller regions with training data coming from specific areas rather than larger and more generic models that try to approach the problem universally, often missing in performance. Finally. the study underscores the importance of public data availability for research purposes as models like the one described in this work would not be easily developed without a large amount of easily and freely accessible data.
\section{Acknowledgements}
This work started as a joke (or challenge) between some fellow postgraduate students and me during our first year (2022) at Eindhoven University of Technology. Coming from Greece, a country well known for its Mediterranean climate and mild winter season, we were impatiently waiting for snow in Eindhoven when the winter started. My friends were awaiting snow every day, sending screenshots from their weather apps every morning in student group chats that only forecasted snow with a probability of 40\% or 50\%. Some friends were still arguing, that a snowstorm was close even when the sun was apparent and bright the whole day and they were claiming so, just because the weather app was forecasting snow with a low probability for the next hour. Then I decided that they needed a data-driven and science-backed response. This is why I would like to acknowledge and thank Michalis Galanis, Giannis Badakis, Stylianos Balomenos, George Spyridakis, Theodoros Chronopoulos and George Tsimpiskakis for posing the challenge. 
\bibliographystyle{unsrt}  
\bibliography{references}

\end{document}